\documentclass[runningheads]{llncs}
\usepackage{graphicx}
\usepackage{epstopdf}
\usepackage{times}
\usepackage{epsfig}
\usepackage{amsmath,amssymb} % define this before the line numbering.
\usepackage{booktabs} 
\usepackage{enumitem}
\usepackage{xcolor}
\usepackage{subfig}
\usepackage{color}
\usepackage{wrapfig}
\usepackage{sidecap}

\usepackage[width=122mm,left=12mm,paperwidth=146mm,height=193mm,top=12mm,paperheight=217mm]{geometry}
\begin{document}
% \linenumbers
\pagestyle{headings}
\mainmatter
\def\ECCV18SubNumber{1403}  % Insert your submission number here

\title{Jointly Discovering Visual Objects and Spoken Words from Raw Sensory Input} % Replace with your title

% Anonymous for submission
%\titlerunning{ECCV-18 submission ID \ECCV18SubNumber}
%\authorrunning{ECCV-18 submission ID \ECCV18SubNumber}
%\author{Anonymous ECCV submission}
%\institute{Paper ID \ECCV18SubNumber}

% Camera Ready
\titlerunning{Jointly Discovering Visual Objects and Spoken Words from Raw Sensory Input}
\authorrunning{Harwath et al.}
\author{David Harwath, Adri\`a Recasens, D\'idac Sur\'is, Galen Chuang,\\Antonio Torralba, and James Glass}
\institute{Massachusetts Institute of Technology}

\maketitle

\begin{abstract}
In this paper, we explore neural network models that learn to associate segments of spoken audio captions with the semantically relevant portions of natural images that they refer to. We demonstrate that these audio-visual associative localizations emerge from network-internal representations learned as a by-product of training to perform an image-audio retrieval task. Our models operate directly on the image pixels and speech waveform, and do not rely on any conventional supervision in the form of labels, segmentations, or alignments between the modalities during training. We perform analysis using the Places 205 and ADE20k datasets demonstrating that our models implicitly learn semantically-coupled object and word detectors.
\end{abstract}

\section{Introduction}
\begin{wrapfigure}{R}{.5\textwidth}
	\vspace{-.25cm}
    \centering
    \includegraphics[width=0.9\linewidth]{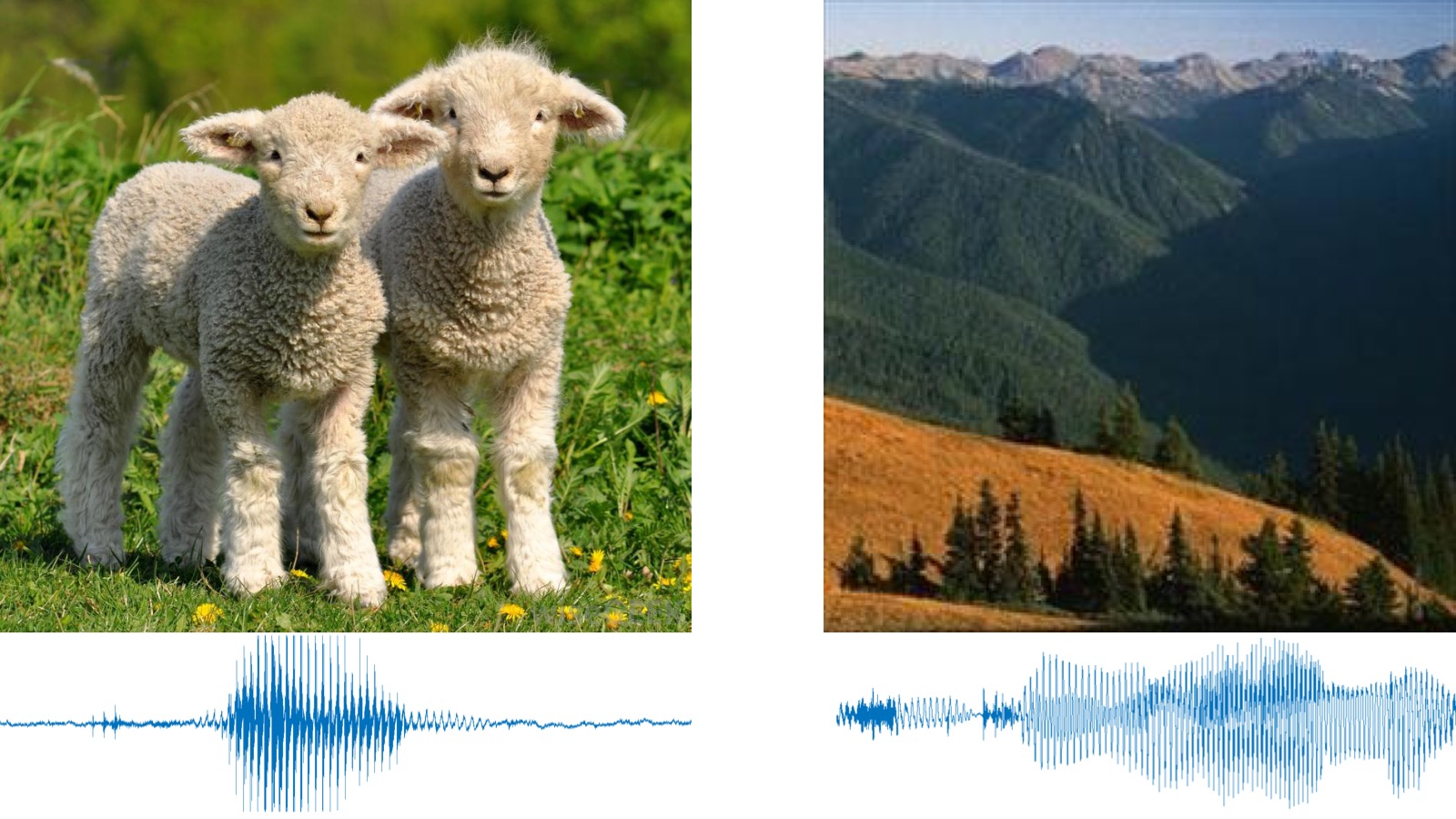}
    \caption{The input to our models: images paired with waveforms of speech audio.}
    \label{fig:images_with_waveforms}
    \vspace{-.75cm}
\end{wrapfigure}

%Humans learn to communicate with speech audio, not text on a screen. They learn to understand speech in an unsupervised fashion, aided not by ground-truth annotations, but by multi-modal context and environmental interaction. In this paper, we do not attempt to model the cognitive development of humans, but instead ask whether a machine can jointly learn spoken language and visual perception when faced with similar constraints.
%\textcolor{red}{TODO: add supporting references to cogsci work on children learning. Also put more emphasis on the completely unsupervised nature of the learning that we are doing in both modalities}
Babies face an impressive learning challenge: they must learn to visually perceive the world around them, and to use language to communicate. They must discover the objects in the world and the words that refer to them. They must solve this problem when both inputs come in raw form: unsegmented, unaligned, and with enormous appearance variability both in the visual domain (due to pose, occlusion, illumination, etc.) and in the acoustic domain (due to the unique voice of every person, speaking rate, emotional state, background noise, accent, pronunciation, etc.). Babies learn to understand speech and recognize objects in an extremely weakly supervised fashion, aided not by ground-truth annotations, but by observation, repetition, multi-modal context, and environmental interaction~\cite{dupoux_2018,SPELKE199029}. In this paper, we do not attempt to model the cognitive development of humans, but instead ask whether a machine can jointly learn spoken language and visual perception when faced with similar constraints; that is, with inputs in the form of unaligned, unannotated raw speech audio and images (Figure \ref{fig:images_with_waveforms}). To that end, we present models capable of jointly discovering words in raw speech audio, objects in raw images, and associating them with one another.

There has recently been a surge of interest in bridging the vision and natural language processing (NLP) communities, in large part thanks to the ability of deep neural networks to effectively model complex relationships within multi-modal data. %These visual-linguistic models have immense potential to address challenging problems within both communities. Language offers a far more flexible and naturalistic way of annotating visual data that goes beyond rigidly defined class labels. It also opens the door for completely new problems, such as caption generation and visual question answering (VQA). Because human language is grounded in the real world, the linguistic representations that can be learned with the benefit of visual context have the potential to be far more semantically rich than text-only models.
Current work  bringing together vision and language~\cite{karpathy_2015,vinyals_2015,fang_2015,xu_2015,densecap,antol_2015,malinowski_2014,malinowski_2015,gao_2015,ren_2015,guess_what,reed_2016} relies on written text. In this situation, the linguistic information is presented in a pre-processed form in which words have been segmented and clustered. The text word {\it car} has no variability between sentences (other than synonyms, capitalization, etc.), and it is already segmented apart from other words. This is dramatically different from how children learn language. The speech signal is continuous, noisy, unsegmented, and exhibits a wide number of non-lexical variabilities. The problem of segmenting and clustering the raw speech signal into discrete words is analogous to the problem of visual object discovery in images - the goal of this paper is to address both problems jointly. 

Recent work has focused on cross modal learning between vision and sounds \cite{Owens16,Owens2016b,NIPS2016_6146,look_listen_learn}. This work has focused on using ambient sounds and video to discover sound generating objects in the world. In our work we will also use both vision and audio modalities except that the audio corresponds to speech. In this case, the problem is more challenging as the portions of the speech signal that refer to objects are shorter, creating a more challenging temporal segmentation problem, and the number of categories is much larger. Using vision and speech was first studied in \cite{harwath_nips}, but it was only used to relate full speech signals and images using a global embedding. Therefore the results focused on image and speech retrieval. Here we introduce a model able to segment both words in speech and objects in images without supervision. 

The premise of this paper is as follows: given an image and a raw speech audio recording describing that image, we propose a neural model which can highlight the relevant regions of the image as they are being described in the speech. What makes our approach unique is the fact that we do not use any form of conventional speech recognition or transcription, nor do we use any conventional object detection or recognition models. In fact, both the speech and images are completely unsegmented, unaligned, and unannotated during training, aside from the assumption that we know which images and spoken captions belong together as illustrated in Figure \ref{fig:images_with_waveforms}. We train our models to perform semantic retrieval at the whole-image and whole-caption level, and demonstrate that detection and localization of both visual objects and spoken words emerges as a by-product of this training.

%%%%%%%%% BODY TEXT
\section{Prior Work}
\textbf{Visual Object Recognition and Discovery.} 
%Classification of visual objects (or other patterns) is a longstanding problem within the computer vision community, with the MNIST~\cite{mnist} handwritten digit task being a classic and widely known example. Recent progress in the field has been driven in part by recurring challenge competitions such as ISLVRC~\cite{ILSVRC15}. Since 2012, the task has been dominated by deep convolutional neural networks (CNNs), popularized by Krizhevsky \etal~\cite{alexnet}. Since that time, improved variants of the basic CNN architecture have continued to push the state of the art \cite{resnet, vgg}. While classification asks the question of ``what'', object detection and localization (also part of the ISLVRC suite of tasks) address the problem of ``where''. 
State of the art systems are trained using bounding box annotations for the training data \cite{girshick_2013,yolo}, however other works investigate weakly-supervised or unsupervised object localization \cite{bergamo_2014,cho_2015,cinbis_2016,zhou_2015}. A large body of research has also focused on unsupervised visual object discovery, in which case there is no labeled training dataset available. One of the first works within this realm is \cite{weber_2000}, which utilized an iterative clustering and classification algorithm to discover object categories. Further works borrowed ideas from textual topic models \cite{russell_2006}, assuming that certain sets of objects generally appear together in the same image scene. More recently, CNNs have been adapted to this task \cite{doersch_2015,guerin_2017}, for example by learning to associate image patches which commonly appear adjacent to one another. 

\textbf{Unsupervised Speech Processing.} Automatic speech recognition (ASR) systems %have a long history and 
have recently made great strides thanks to the revival of deep neural networks. 
%The technology is now close to reaching human parity within certain language and domains \cite{xiong_2016}. However, this performance comes at an enormous cost. 
Training a state-of-the-art ASR system requires thousands of hours of transcribed speech audio, along with expert-crafted pronunciation lexicons and text corpora covering millions, if not billions of words for language model training. The reliance on expensive, highly supervised training paradigms has restricted the application of ASR to the major languages of the world, accounting for a small fraction of the more than 7,000 human languages spoken worldwide ~\cite{ethnologue}. Within the speech community, there is a continuing effort to develop algorithms less reliant on transcription and other forms of supervision. Generally, these take the form of segmentation and clustering algorithms whose goal is to divide a collection of spoken utterances at the boundaries of phones or words, and then group together segments which capture the same underlying unit. Popular approaches are based on dynamic time warping~\cite{park_glass_sdtw,jansen_2010,jansen_2011}, or Bayesian generative models of the speech signal~\cite{lee_glass_2012,but_2016,kamper_2016}. Neural networks have thus far been mostly utilized in this realm for learning frame-level acoustic features~\cite{zhang_2012,renshaw_2015,kamper_2015,thiolliere_2015}.

\textbf{Fusion of Vision and Language.} Joint modeling of images and natural language text has gained rapidly in popularity, encompassing tasks such as image captioning~\cite{karpathy_2015,vinyals_2015,fang_2015,xu_2015,densecap}, visual question answering (VQA)~\cite{antol_2015,malinowski_2014,malinowski_2015,gao_2015,ren_2015}, multimodal dialog~\cite{guess_what}, and text-to-image generation~\cite{reed_2016}. While most work has focused on representing natural language with text, there are a growing number of papers attempting to learn directly from the speech signal. A major early effort in this vein was the work of Roy~\cite{roy_2002,roy_2003}, who learned correspondences between images of objects and the outputs of a supervised phoneme recognizer. Recently, it was demonstrated by Harwath et al ~\cite{harwath_nips} that semantic correspondences could be learned between images and speech waveforms at the signal level, with subsequent works providing evidence that linguistic units approximating phonemes and words are implicitly learned by these models~\cite{harwath_acl_2017,drexler_2017,chrupala_2017,alishahi_2017,kamper_2017}. This paper follows in the same line of research, introducing the idea of ``matchmap'' networks which are capable of directly inferring semantic alignments between acoustic frames and image pixels.

\textbf{Fusion of Vision and Sounds.} 
A number of recent models have focused on integrating other acoustic signals to perform unsupervised discovery of objects and ambient sounds~\cite{Owens16,Owens2016b,NIPS2016_6146,look_listen_learn}. Our work concentrates on speech and word discovery. But combining both types of signals (speech and ambient sounds) opens a number of opportunities for future research beyond the scope of this paper.

\section{Spoken Captions Dataset}

\begin{wrapfigure}{R}{.5\textwidth}
\centering
\vspace{-1cm}
\subfloat[]{\includegraphics[width=0.49\linewidth]{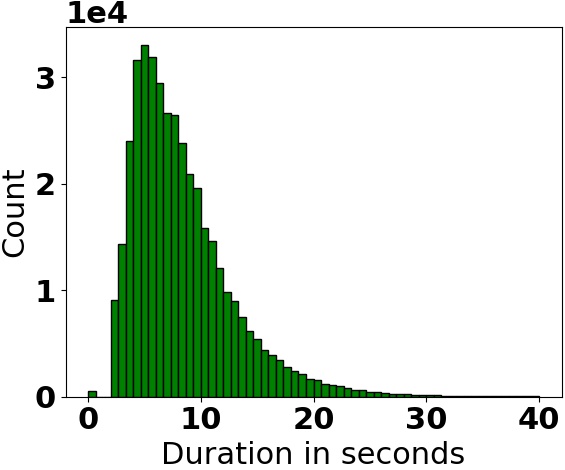}}
\subfloat[]{\includegraphics[width=0.49\linewidth]{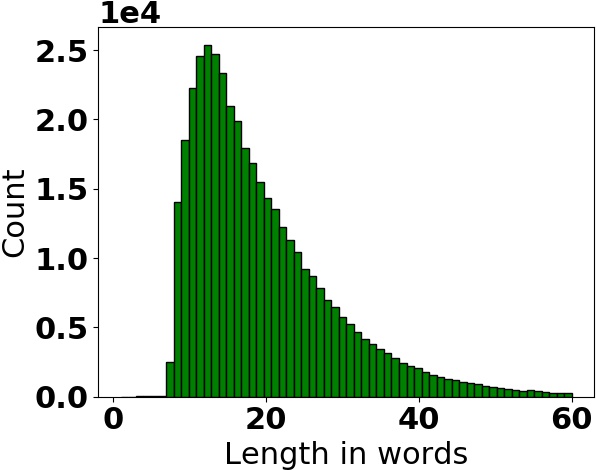}} \\
\subfloat[]{\includegraphics[width=0.49\linewidth]
{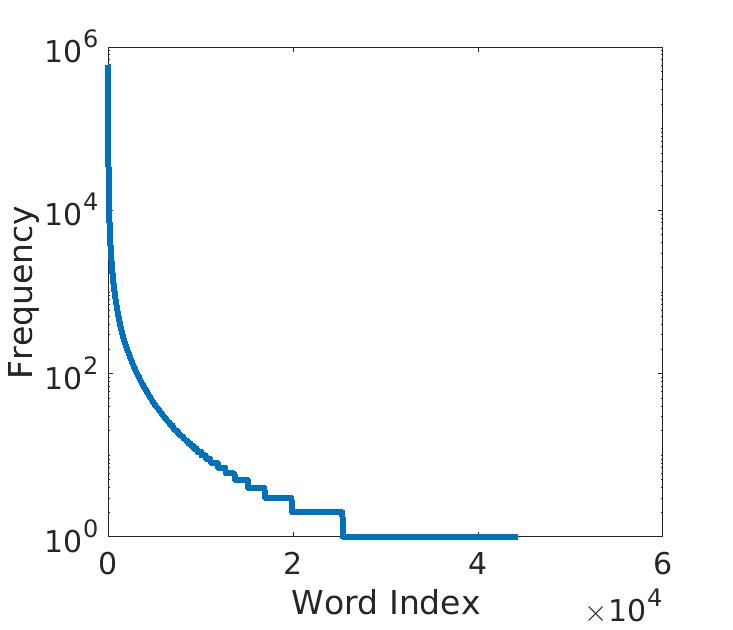}} 
\subfloat[]{\includegraphics[width=0.49\linewidth]
{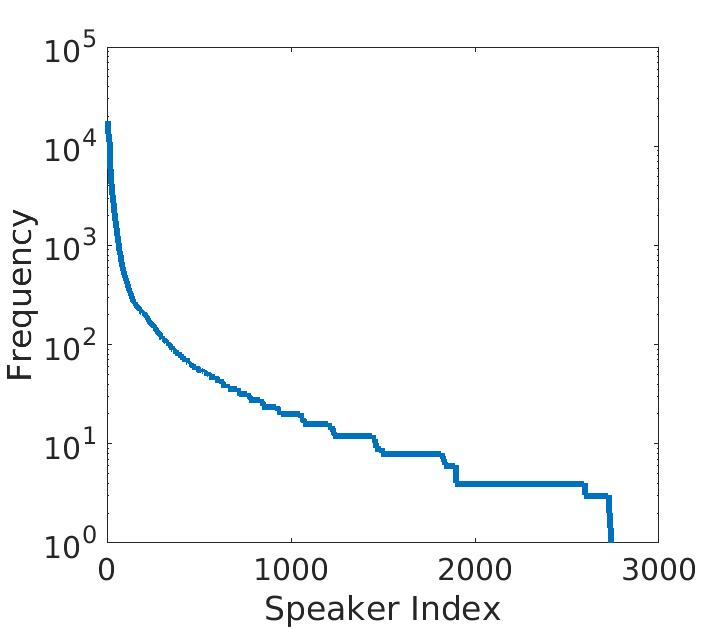}}
\caption[caption]{Statistics of the 400k spoken captions. From left to right, the plots represent (a) the histogram over caption durations in seconds, (b) the histogram over caption lengths in words, (c) the estimated word frequencies across the captions, and (d) the number of captions per speaker.}
\label{fig:data_stats}
\end{wrapfigure}

For training our models, we use the Places Audio Caption dataset~\cite{harwath_nips,harwath_acl_2017}. This dataset contains approximately 200,000 recordings collected via Amazon Mechanical Turk of people verbally describing the content of images from the Places 205~\cite{places} image dataset. We augment this dataset by collecting an additional 200,000 captions, resulting in a grand total of 402,385 image/caption pairs for training and a held-out set of 1,000 additional pairs for validation. In order to perform a fine-grained analysis of our models ability to localize objects and words, we collected an additional set of captions for 9,895 images from the ADE20k dataset~\cite{ade20k} whose underlying scene category was found in the Places 205 label set. The ADE20k data contains pixel-level object labels, and when combined with acoustic frame-level ASR hypotheses, we are able to determine which underlying words match which underlying objects. In all cases, we follow the original Places audio caption dataset and collect 1 caption per image. Aggregate statistics over the data are shown in Figure \ref{fig:data_stats}. While we do not have exact ground truth transcriptions for the spoken captions, we use the Google ASR engine to derive hypotheses which we use for experimental analysis (but not training, except in the case of the text-based models). A vocabulary of 44,342 unique words were recognized within all 400k captions, which were spoken by 2,683 unique speakers. The distributions over both words and speakers follow a power law with a long tail (Figure ~\ref{fig:data_stats}). We also note that the free-form nature of the spoken captions generally results in longer, more descriptive captions than exist in text captioning datasets. While MSCOCO \cite{mscoco} contains an average of just over 10 words per caption, the places audio captions are on average 20 words long, with an average duration of 10 seconds. The extended Places 205 audio caption corpus, the ADE20k caption data, and a PyTorch implementation of the model training code will be made available at (URL hidden).
% places 205 vocab size = 43,953
% ade20k vocab size = 6,096
% places 205 + ade20k vocab size = 44,342
% number of speakers in places 205 = 2683
% number of speakers in ade20k = 72
% number of unique speakers across both places 205 and ade20k = 2,735

\section{Models}
\begin{figure*}[h]
    \centering
    \includegraphics[width=1\linewidth]{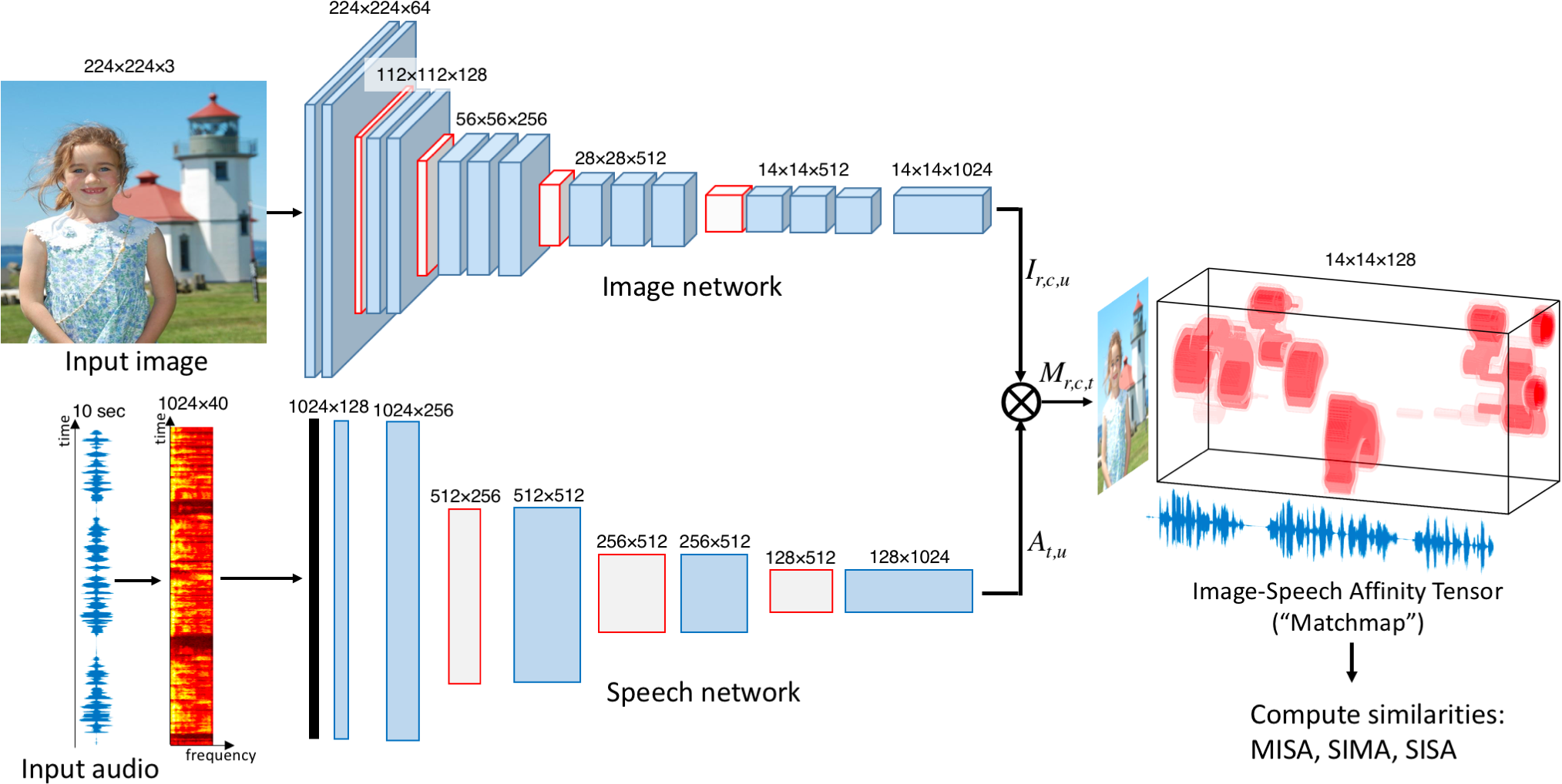}
    \caption{The audio-visual matchmap model architecture (left), along with an example matchmap output (right), displaying a 3-D density of spatio-temporal similarity. Conv layers shown in blue, pooling layers shown in red, and BatchNorm layer shown in black. Each conv layer is followed by a ReLU. The first conv layer of the audio network uses filters that are 1 frame wide and span the entire frequency axis; subsequent layers of the audio network are hence 1-D convolutions with respective widths of 11, 17, 17, and 17. All maxpool operations in the audio network are 1-D along the time axis with a width of 3. An example spectrogram input of approx. 10 seconds (1024 frames) is shown to illustrate the pooling ratios.}
    \label{fig:matchmap_model}
\end{figure*}

Our model is similar to that of Harwath et al ~\cite{harwath_nips}, in which a pair of convolutional neural networks (CNN)~\cite{mnist} are used to independently encode a visual image and a spoken audio caption into a shared embedding space. What differentiates our models from prior work is the fact that instead of mapping entire images and spoken utterances to fixed points in an embedding space, we learn representations that are \textit{distributed} both spatially and temporally, enabling our models to directly co-localize within both modalities. Our models are trained to optimize a ranking-based criterion~\cite{bromley_1994,karpathy_2014,harwath_nips}, such that images and captions that belong together are more similar in the embedding space than mismatched image/caption pairs. Specifically, across a batch of $B$ image/caption pairs $(I_j, A_j)$ (where $I_j$ represents the output of the image branch of the network for the $j^{th}$ image, and $A_j$ the output of the audio branch for the $j^{th}$ caption) we compute the loss as
\begin{equation}
\begin{split}
\text{L} = \sum_{j=1}^B & \Big(\max(0, S(I_j, A_j^{imp}) - S(I_j, A_j) + \eta) \\ &+ \max(0, S(I_j^{imp}, A_j) - S(I_j, A_j) + \eta) \Big),
\label{eq:sampled_margin_ranking_objective}
\end{split}
\end{equation}
where $S(I, A)$ represents the similarity score between an image $I$ and audio caption $A$, $I_j^{imp}$ represents the $j^{th}$ randomly chosen imposter image, $A_j^{imp}$ the $j^{th}$ imposter caption, and $\eta$ is a margin hyperparameter. We sample the imposter image and caption for each pair from the same minibatch, and fix $\eta$ to 1 in our experiments. The choice of similarity function is flexible, which we explore in Section \ref{sec:joining}. This criterion directly enables semantic retrieval of images from captions and vice versa, but in this paper our focus is to explore how object and word \textit{localization} naturally emerges as a by-product of this training scheme. An illustration of our two-branch matchmap networks is shown in Figure \ref{fig:matchmap_model}. 
%In the remainder of this section, 
Next, we describe the modeling for each input mode.

\subsection{Image Modeling}
We follow~\cite{harwath_nips,harwath_acl_2017,gelderloos_2016,chrupala_2017,alishahi_2017,kamper_2017} by utilizing the architecture of the VGG16 network~\cite{vgg} to form the basis of the image branch. In all of these prior works, however, the weights of the VGG network were pre-trained on ImageNet, and thus had a significant amount of visual discriminative ability built-in to their models. We show that our models do not require this pre-training, and can be trained end-to-end in a completely unsupervised fashion. Additionally in these prior works, the entire VGG network below the classification layer was utilized to derive a single, global image embedding. One problem with this approach is that coupling the output of \texttt{conv5} to \texttt{fc1} involves a flattening operation, which makes it difficult to recover associations between any neuron above \texttt{conv5} and the spatially localized stimulus which was responsible for its output. We address this issue here by retaining only the convolutional banks up through \texttt{conv5} from the VGG network, and discarding \texttt{pool5} and everything above it. For a 224 by 224 pixel input image, the output of this portion of the network would be a 14 by 14 feature map across 512 channels, with each location within the map possessing a receptive field that can be related directly back to the input. In order to map an image into the shared embedding space, we apply a 3 by 3, 1024 channel, linear convolution (no nonlinearity) to the \texttt{conv5} feature map. Image pre-processing consists of resizing the smallest dimension to 256 pixels, taking a random 224 by 224 crop (the center crop is taken for validation), and normalizing the pixels according to a global mean and variance.  

\subsection{Audio Caption Modeling}
To model the spoken audio captions, we use a model similar to that of ~\cite{harwath_acl_2017}, but modified to output a feature map across the audio during training, rather than a single embedding vector. The audio waveforms are represented as log Mel filter bank spectrograms. Computing these involves first removing the DC component of each recording via mean subtraction, followed by pre-emphasis filtering. The short-time Fourier transform is then computed using a 25 ms Hamming window with a 10 ms shift. We take the squared magnitude spectrum of each frame and compute the log energies within each of 40 Mel filter bands. We treat these final spectrograms as 1-channel images, and model them with the CNN displayed in Figure \ref{fig:matchmap_model}. ~\cite{harwath_nips} utilized truncation and zero-padding of each spectrogram to a fixed length. While this enables batched inputs to the model, it introduces a degree of undesirable bias into the learned representations. Instead, we pad to a length long enough to fully capture the longest caption within a batch, and truncate the output feature map of each caption on an individual basis to remove the frames corresponding to zero-padding. Rather than manually normalizing the spectrograms, we employ a BatchNorm~\cite{batchnorm} layer at the front of the network. Next, we discuss methods for relating the visual and auditory feature maps to one another.

\subsection{Joining the Image and Audio Branches}
\label{sec:joining}
Zhou et al ~\cite{zhou2015cnnlocalization} demonstrate that global average pooling applied to the \texttt{conv5} layer of several popular CNN architectures not only provides good accuracy for image classification tasks, but also enables the recovery of spatial activation maps for a given target class at the \texttt{conv5} layer, which can then be used for object localization. The idea that a pooled representation over an entire input used for training can then be unpooled for localized analysis is powerful because it does not require localized annotation of the training data, or even any explicit mechanism for localization in the objective function or network itself, beyond what already exists in the form of convolutional receptive fields. Although our models perform a ranking task and not classification, we can apply similar ideas to both the image and speech feature maps in order to compute their pairwise similarity, in the hopes to recover localizations of objects and words. Let $I$ represent the output feature map output of the image network branch, $A$ be the output feature map of the audio network branch, and $I^p$ and $A^p$ be their globally average-pooled counterparts. One straightforward choice of similarity function is the dot product between the pooled embeddings, $S(I, A) = I^{pT} A^p$. Notice that this is in fact equivalent to first computing a 3rd order tensor $M$ such that $M_{r,c,t} = I_{r, c, :}^{T} A_{t, :}$, and then computing the average of all elements of $M$. Here we use the colon ($:$) to indicate selection of all elements across an indexing plane; in other words, $I_{r, c, :}$ is a 1024-dimensional vector representing the $(r,c)$ coordinate of the image feature map, and $A_{t, :}$ is a 1024-dimensional vector representing the $t^{th}$ frame of the audio feature map. In this regard, the similarity between the global average pooled image and audio representations is simply the average similarity between \textit{all} audio frames and \textit{all} image regions. We call this similarity scoring function SISA (sum image, sum audio):
\begin{equation}
\text{SISA}(M) = \frac{1}{N_r N_c N_t}\sum_{r=1}^{N_r}\sum_{c=1}^{N_c}\sum_{t=1}^{N_t}{M_{r,c,t}}
\end{equation}
Because $M$ reflects the localized similarity between a small image region (possibly containing an object) and a small segment of audio (possibly containing a word), we dub $M$ the ``matchmap'' tensor between and image and an audio caption. As it is not completely realistic to expect all words within a caption to simultaneously match all objects within an image, we consider computing the similarity between an image and an audio caption using several alternative functions of the matchmap density. By replacing the averaging summation over image patches with a simple maximum, MISA (max image, sum audio) effectively matches each frame of the caption with the most similar image patch, and then averages over the caption frames:
\begin{equation}
\text{MISA}(M) = \frac{1}{N_t}\sum_{t=1}^{N_t}{\max_{r,c}(M_{r,c,t})}
\end{equation}
By preserving the sum over image regions but taking the maximum across the audio caption, SIMA (sum image, max audio) matches each image region with only the audio frame with the highest similarity to that region:
\begin{equation}
\text{SIMA}(M) = \frac{1}{N_r N_c}\sum_{r=1}^{N_r}{\sum_{c=1}^{N_c}{\max_t(M_{r,c,t})}}
\end{equation}
In the next section, we explore the use of these similarities for learning semantic correspondences between objects within images and spoken words within their captions.

\section{Experiments}
\subsection{Image and Caption Retrieval}
\label{sec:retrieval}

All models were trained using the sampled margin ranking objective outlined in Equation \ref{eq:sampled_margin_ranking_objective}, using stochastic gradient descent with a batch size of 128. We used a fixed momentum of 0.9 and an initial learning rate of 0.001 that decayed by a factor of 10 every 70 epochs; generally our models converged in less than 150 epochs. We use a held-out set of 1,000 image/caption pairs from the Places audio caption dataset to validate the models on the image/caption retrieval task, similar to the one described in ~\cite{harwath_nips,harwath_acl_2017,chrupala_2017,alishahi_2017}. This task serves to provide a single, high-level metric which captures how well the model has learned to semantically bridge the audio and visual modalities. While providing a good indication of a model's overall ability, it does not directly examine which specific aspects of language and visual perception are being captured. Table \ref{tab:retrieval_algos} displays the image/caption recall scores achieved when training a matchmap model using the SISA, MISA, and SIMA similarity functions, both with a fully randomly initialized network as well as with an image branch pre-trained on ImageNet. In all cases, the MISA similarity measure is the best performing, although all three measures achieve respectable scores. Unsurprisingly, using a pre-trained image network significantly increases the recall scores. In Table \ref{tab:retrieval_algos}, we compare our models against reimplementations of two previously published speech-to-image models (both of which utilized pre-trained VGG16 networks). We also compare against baselines that operate on automatic speech recognition (ASR) derived text transcriptions of the spoken captions. The text-based model we used is based on the architecture of the speech and image model, but replaces the speech audio branch with a CNN that operates on word sequences. The ASR text network uses a 200-dimensional word embedding layer, followed by a 512 channel, 1-dimensional convolution across windows of 3 words with a ReLU nonlinearity. A final convolution with a window size of 3 and no nonlinearity maps these activations into the 1024 multimodal embedding space. Both previously published baselines we compare to used the full VGG network, deriving an embedding for the entire image from the \texttt{fc2} outputs. In the pre-trained case, our best recall scores for the MISA model outperform \cite{harwath_nips} overall as well as \cite{harwath_acl_2017} on image recall; the caption recall score is slightly lower than that of \cite{harwath_acl_2017}. This demonstrates that there is not much to be lost when doing away with the fully connected layers of VGG, and much to be gained in the form of the localization matchmaps.
\begin{table}[htb]
\setlength\tabcolsep{2pt}
\small
  \caption{Recall scores on the held out set of 1,000 images/captions for the three matchmap similarity functions. We also show results for the baseline models which use automatic speech recognition-derived text captions. The (P) indicates the use of an image branch pre-trained on ImageNet.}
  \label{tab:retrieval_algos}
  \centering
  \begin{tabular}{ccccccccccccc}
    \toprule
    \multicolumn{1}{c}{} & \multicolumn{6}{c}{Speech} & \multicolumn{6}{c}{ASR Text} \\
    \multicolumn{1}{c}{} & \multicolumn{3}{c}{Caption to Image} & \multicolumn{3}{c}{Image to Caption}& \multicolumn{3}{c}{Caption to Image} & \multicolumn{3}{c}{Image to Caption}\\
    Model & R@1 & R@5 & R@10 & R@1 & R@5 & R@10 & R@1 & R@5 & R@10 & R@1 & R@5 & R@10\\
    \cmidrule(lr){2-4}\cmidrule(lr){5-7}\cmidrule(lr){8-10}\cmidrule(lr){11-13}
    SISA & .063 & .191 & .274 & .048 & .166 & .249 & .136 & .365 & .503 & .106 & .309 & .430 \\
    MISA & .079 & .225 & .314 & .057 & .191 & .291 & .162 & .417 & .547 & .113 & .309 & .447 \\
    SIMA & .073 & .213 & .284 & .065 & .168 & .255 & .134 & .389 & .513 & .145 & .336 & .459 \\
    \midrule
    SISA(P) & .165 & .431 & .559 & .120 & .363 & .506 & .230 & .525 & .665 & .174 & .462 & .611 \\
    MISA(P) & .200 & .469 & .604 & .127 & .375 & .528 & .271 & .567 & .701 & .183 & .489 & .622 \\
    SIMA(P) & .147 & .375 & .506 & .139 & .367 & .483 & .215 & .518 & .639 & .220 & .494 & .599 \\
    \midrule
    \cite{harwath_nips}(P)& .148 & .403 & .548 & .121 & .335 & .463 & - & - & - &- &- &- \\
    \cite{harwath_acl_2017}(P)& .161 & .404 & .564 & .130 & .378 &  .542 & - & - & - &- &- &-\\
    \bottomrule
  \end{tabular}
\end{table}

\begin{figure*}
	\includegraphics[width=.48\linewidth]{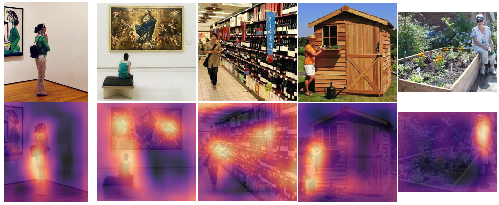}
    \includegraphics[width=.48\linewidth]{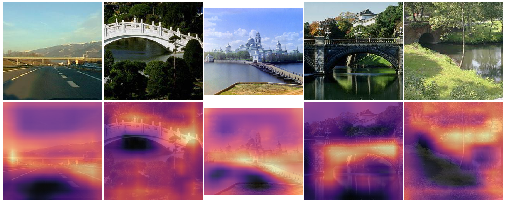}
    \includegraphics[width=.48\linewidth]{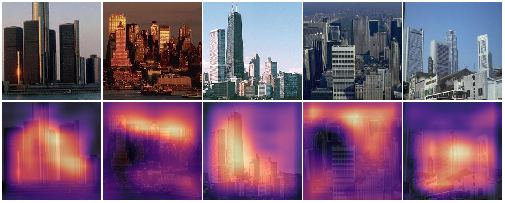}
    \includegraphics[width=.48\linewidth]{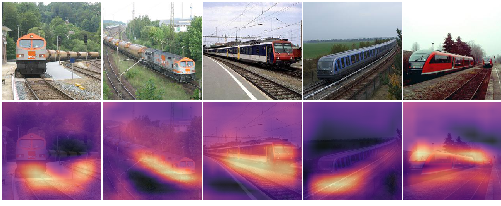}
    \includegraphics[width=.48\linewidth]{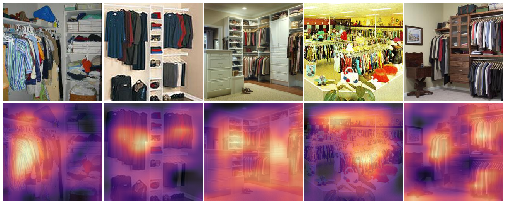}
    ~~~~~~~~\includegraphics[width=.48\linewidth]{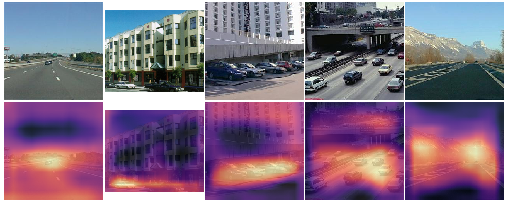}
%\vspace{-.3in}
	\caption{Speech-prompted localization maps for several word/object pairs. From top to bottom and from left to right, the queries are instances of the spoken words ``WOMAN,'' ``BRIDGE,'', ``SKYLINE'', ``TRAIN'', ``CLOTHES'' and ``VEHICLES'' extracted from each image's accompanying speech caption.}
	\label{fig:localization_maps}
\end{figure*}

\subsection{Speech-Prompted Object Localization.}
To evaluate our models' ability to associate spoken words with visual objects in a more fine-grained sense, we use the spoken captions for the ADE20k~\cite{ade20k} dataset. The ADE20k images contain pixel-level object masks and labels - in conjunction with a time-aligned transcription produced via ASR (we use the public Google SpeechRecognition API for this purpose), we can associate each matchmap cell with a specific visual object label as well as a word label. These labels enable us to analyze which words are being associated with which objects. We do this by performing speech-prompted object localization. Given a word in the speech beginning at time $t_1$ and ending at time $t_2$, we derive a heatmap across the image by summing the matchmap between $t_1$ and $t_2$. We then normalize the heatmap to sit within the interval [0,1], threshold the heatmap, and evaluate the intersection over union (IoU) of the detection mask with the ADE20k label mask for whatever object was referenced by the word.

Because there are a very large number of different words appearing in the speech, and no one-to-one mapping between words and ADE20k objects exists, we manually define a set of 100 word-object pairings. We choose commonly occurring (at least 9 occurrences) pairs that are unambiguous, such as the word ``building'' and object ``building,'' the word ``man'' and the ``person'' object, etc. For each word-object pair, we compute an average IoU score across all instances of the word-object pair appearing together in an ADE20k image and its associated caption. We then average these scores across all 100 word-object pairs and report results for each model type in Table \ref{tab:IoU_scores}. We also report the IoU scores for the ASR text-based baseline models described in Section \ref{sec:retrieval}. Figure \ref{fig:localization_maps} displays a sampling of localization heatmaps for several query words using the non-pretrained speech MISA network.

\begin{wraptable}{R}{.5\textwidth}
\centering
  \setlength\tabcolsep{3pt}
  \small
  \vspace{-.75cm}
  \caption{Speech-prompted and ASR-prompted object localization IoU scores on the ADE20k data, averaged across the 100 handpicked word-object pairs. `Rand.' stands for a fully randomly initialized network, while `Pre.' indicates that the image branch of the model was initialized with VGG16 weights from ImageNet.}
    \label{tab:IoU_scores}
  \begin{tabular}{ccccc}
    \toprule
    \multicolumn{1}{c}{} & \multicolumn{2}{c}{Speech} & \multicolumn{2}{c}{ASR Text} \\
	Sim. Func. & Rand. & Pre. & Rand. & Pre. \\
    \cmidrule(lr){2-3}\cmidrule(lr){4-5}
    \midrule
    AVG & .1637 & .1970 & .1750 & .2161 \\
    MISA & .1795 & .2324 & .2060 & .2413 \\
    SIMA & .1607 & .1857 & .1743 & .1995 \\
    \bottomrule
  \end{tabular}
\end{wraptable}

\subsection{Clustering of Audio-Visual Patterns}
\begin{figure}
    \includegraphics[width=1\linewidth]{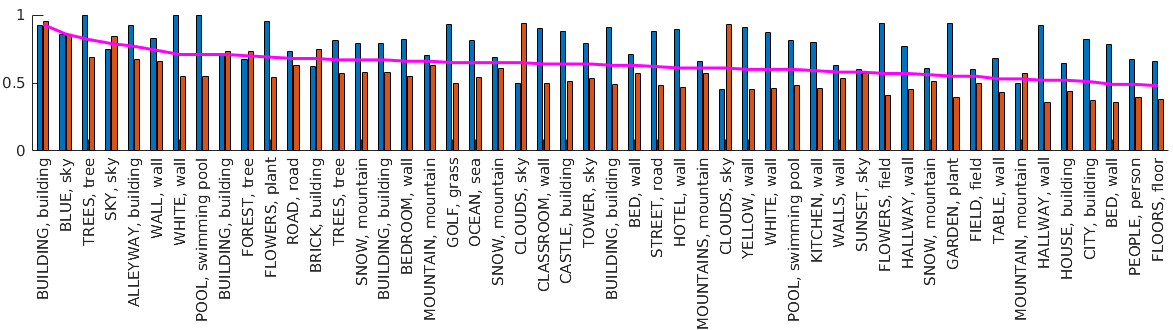}
%\vspace{-.3in}
    \caption{Some clusters (speech and visual) found by our approach. Each cluster is jointly labeled with the most common word (capital letters) and object (lowercase letters). For each cluster we show the precision for both the word (blue) and object (red) labels, as well as their harmonic mean (magenta). The average cluster size across the top 50 clusters was 44.}
    \label{fig:clusters}
\end{figure}

The next experiment we consider is automatic discovery of audio-visual clusters from the ADE20k matchmaps using the fully random speech MISA network. Once a matchmap has been computed for an image and caption pair, we smooth it with an average or max pooling window of size 7 across the temporal dimension before binarizing it according to a threshold. In practice, we set this threshold on a matchmap-specific basis to be 1.5 standard deviations above the mean value of the smoothed matchmap. Next, we extract volumetric connected components and their associated masks over the image and audio. We average pool the image and audio feature maps within these masks, producing a pair of vectors for each component. Because we found the image and speech representations to exhibit different dynamic ranges, we first rescale them by the average L2 norms across all derived image vectors and speech vectors, respectively. We concatenate the image and speech vectors for each component, and finally perform Birch clustering~\cite{birch} with 1000 target clusters for the first step, and an agglomerative final step that resulted in 135 clusters. To derive word labels for each cluster, we take the most frequent word label as overlapped by the components belonging to a cluster. To generate the object labels, we compute the number of pixels belonging to each ADE20k class assigned to a particular cluster, and take the most common label. We display the labels and their purities for the top 50 most pure clusters in Figure \ref{fig:clusters}.

\subsection{Concept discovery: building an image-word dictionary}
Figure \ref{fig:clusters} shows the clusters learned by our model. Interestingly, the audio and image networks are able to agree to a common representation of knowledge, clustering similar concepts together. Since both representations are directly multiplied by a dot product, both networks have to agree on the meaning of these different dimensions. To further explore this phenomenon, we decided to visualize the concepts associated with each of these dimensions for both image and audio networks separately and then find a quantitative strategy to evaluate the agreement.

\begin{figure*}
    \includegraphics[width=1\linewidth]{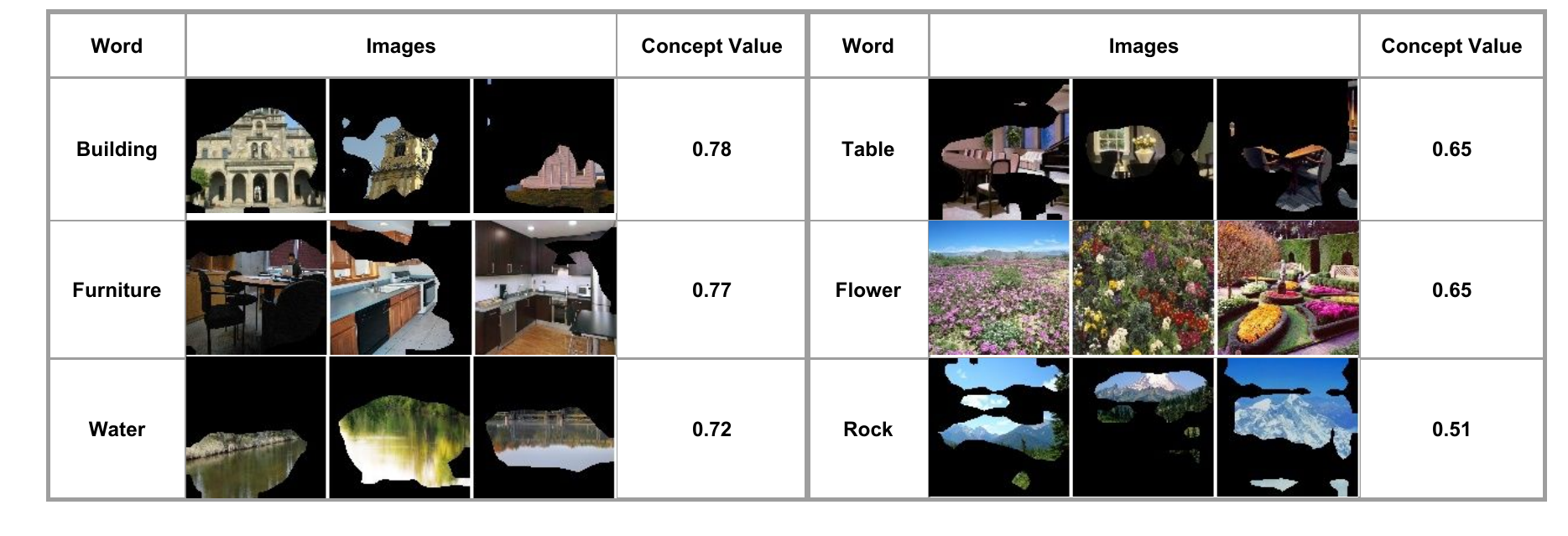}
\vspace{-.3in}
\caption{Matching the most activated images in the image network and the activated words in the audio network we can establish pairs of image-word, as shown in the figure. We also define a concept value, which captures the agreement between both networks and ranges from 0 (no agreement) to 1 (full agreement).}
    \label{fig:word_images}
\end{figure*}

To visualize the concepts associated with each of the dimensions in the image path, we use the unit visualization technique introduced in \cite{zhou2014object}. A set of images is run through the image network and the ones that activate the most that particular dimension get selected. Then, we can visualize the spatial activations in the top activated images. The same procedure can be done for the audio network, where we get a set of descriptions that maximally activate that neuron. Finally, with the temporal map, we can find which part of the description has produced that activation. Some most activated words and images can be found in Figure \ref{fig:word_images}. We show four dimensions with their associated most activated word in the audio neuron, and the most activated images in the image neuron. Interestingly, these pairs of concepts have been found completely independently, as we did not use the final activation (after the dot product) to pick the images. 

The pairs image-word allow us to explore multiple questions. First, can we build an image-word dictionary by only listening to descriptions of images? As we show in Figure \ref{fig:word_images}, we do. It is important to remember that these pairs are learned in a completely unsupervised fashion, without any concept previously learned by the network. Furthermore, in the scenario of a language without written representation, we could just have an image-audio dictionary using exactly the same technique.

Another important question is whether we quantify the quality of the network using this audio-visual dictionary. It is expected that the quality of the dictionary is related with the quality of the network: the better the concepts are learned, the best the network performs. In this section we propose a metric to quantify this dictionary quality. This metric will help us to compute the quality of each individual neuron and the quality of one particular model. 

To quantify the quality of the dictionary, we need to find a common space between the written descriptions and the image activations. Again, this common space comes from a segmentation dataset. Using \cite{ade20k}, we can rank the most detected objects by each of the neurons. We pass through the network approx. 10,000 images from the ADE20k dataset and check for each neuron which classes are most activated for that particular dimension. As a result, we have a set of object labels associated with the image neuron (coming from the segmentation classes), and a word associated with the audio neuron. Using the WordNet tree, we can compute the word distance between these concepts and define the following metric:

\begin{equation}
c = \sum_{i=1}^{|O^{\text{im}}|}{w_i{Sim}_{\text{wup}}(o_i^{\text{im}},o^{\text{au}})},
\end{equation}

with $o_i^{\text{im}}\in O^{\text{im}}$, where $O^{\text{im}}$ is the set of classes present in the TOP5 segmented images and ${Sim}_{\text{wup}}(.,.)$ is the Wu and Palmer WordNet-based similarity, with range [0,1] (higher is more similar). We weight the similarity with $w_i$, which is proportional to intersection over union of the pixels for that class into the masked region of the image. Using this metric, we can then assign one value per dimension, which measures how well both the audio network and the image network agree on that particular concept. The numerical values for six concept pairs are shown in Figure \ref{fig:word_images}. We see how neurons with higher value are cleaner and more related with its counterpart. The bottom right neuron shows an example of low concept value, where the audio word is ``rock'' but the neuron images show mountains in general. Anecdotally, we found $c>0.6$ to be a good indicator that a concept has been learned.

\begin{wrapfigure}{R}{.52\textwidth}
    \centering
     \raisebox{0pt}[\dimexpr\height-2\baselineskip\relax]{\includegraphics[width=0.95\linewidth]{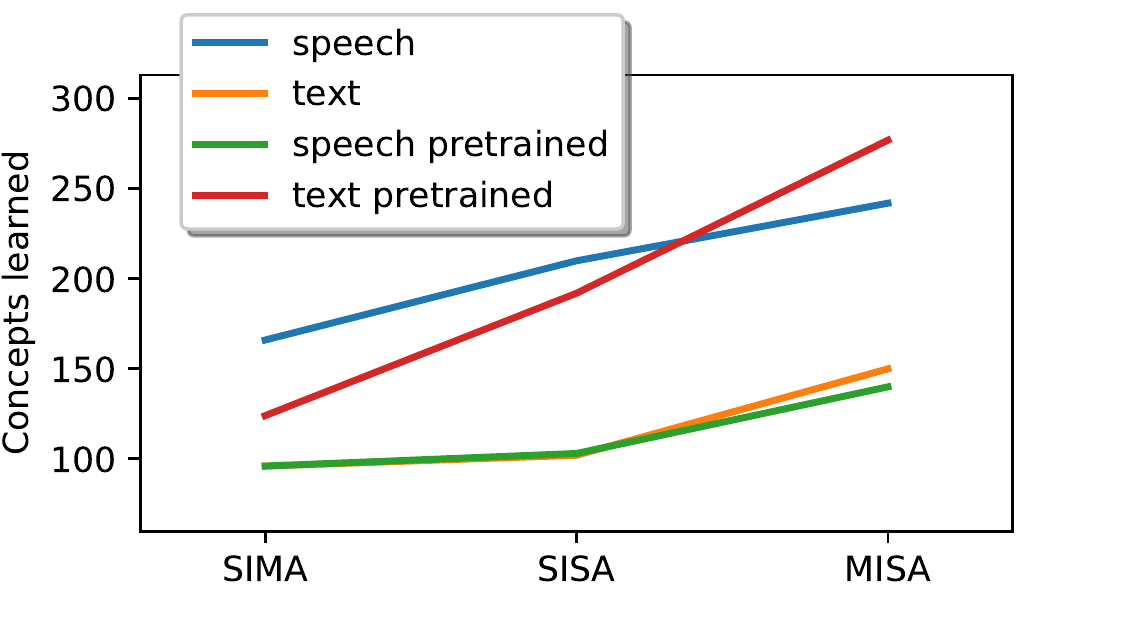}}
    \caption{We study the number of concepts learned by the different networks with different losses, and we find it is consistently lower for SIMA and higher for MISA.}
     % \vspace{-1cm}
    \label{fig:concept_loss}
\end{wrapfigure}

% \begin{SCfigure}
% \includegraphics[width=0.5\linewidth]{data_images/concepts.pdf}
% \caption{We study the number of concepts learned by the different networks with different losses. We find that the number of concepts is consistently lower for SIMA and higher for MISA.}
% \label{fig:concept_loss}
% \end{SCfigure}

Finally, we analyze the relation between the concepts learned and the architecture used. In Figure \ref{fig:concept_loss}, we show for the three different losses we tried, the number of concepts learned by the four networks. Interestingly, the four maintain the same order in the three different cases, indicating that the architecture does influence the number of concepts learned.

% Given the observed correlation between network performance and concepts learned, we also studied the evolution of the concepts during the training of the network... \textcolor{red}{figure of table...? (first, compute results and see what do we have)}

\subsection{Matchmap Visualizations and Videos}
We can visualize the matchmaps in several ways. The 3-dimensional density shown in Figure \ref{fig:matchmap_model} is perhaps the simplest, although it can be difficult to read as a still image. Instead, we can treat it as a stack of masks overlayed on top of the image and played back as a video. We use the matchmap score to modulate the alpha channel of the image across time, and play back the resulting video at 12.5 frames per second so that it temporally aligns with the speech audio playback. The resulting video is able to highlight the salient regions of the images as the speaker is describing them. In the supplementary materials, we include a large number of these videos. 

\begin{figure}
    \includegraphics[width=1\linewidth]{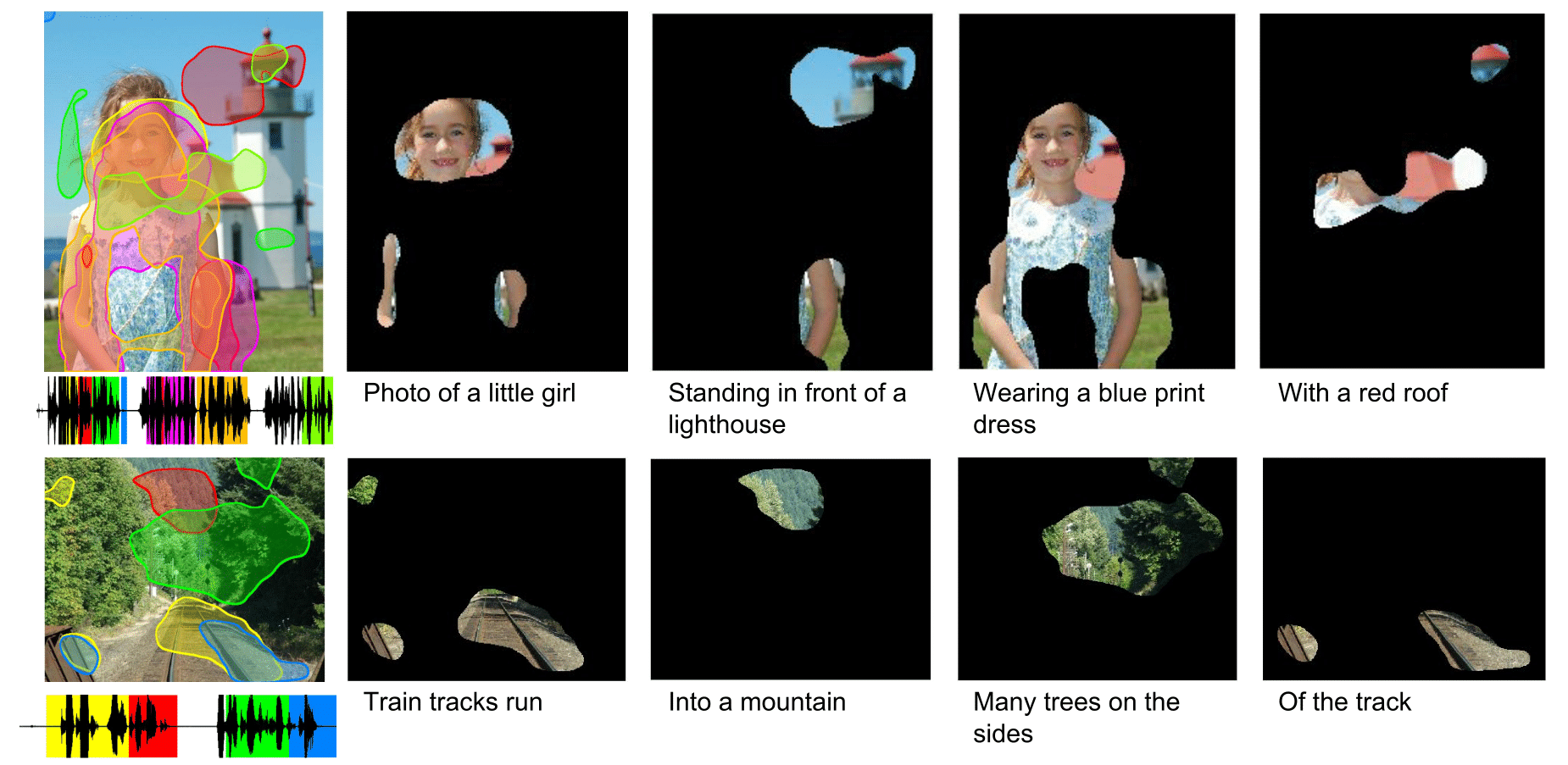}
\vspace{-.2in}
    \caption{On the left are shown two images and their speech signals. Each color corresponds to one connected component derived from two matchmaps from a fully random MISA network. The masked images on the right display the segments that correspond to each piece of the speech signal. For clarity, we show the caption words obtained from the ASR transcriptions below the masks. Note that those words were never used for learning, only for analysis. See full video on supplementary material.}
    \label{fig:matchmapvis}
\end{figure}

\begin{figure}
    \includegraphics[width=1\linewidth]{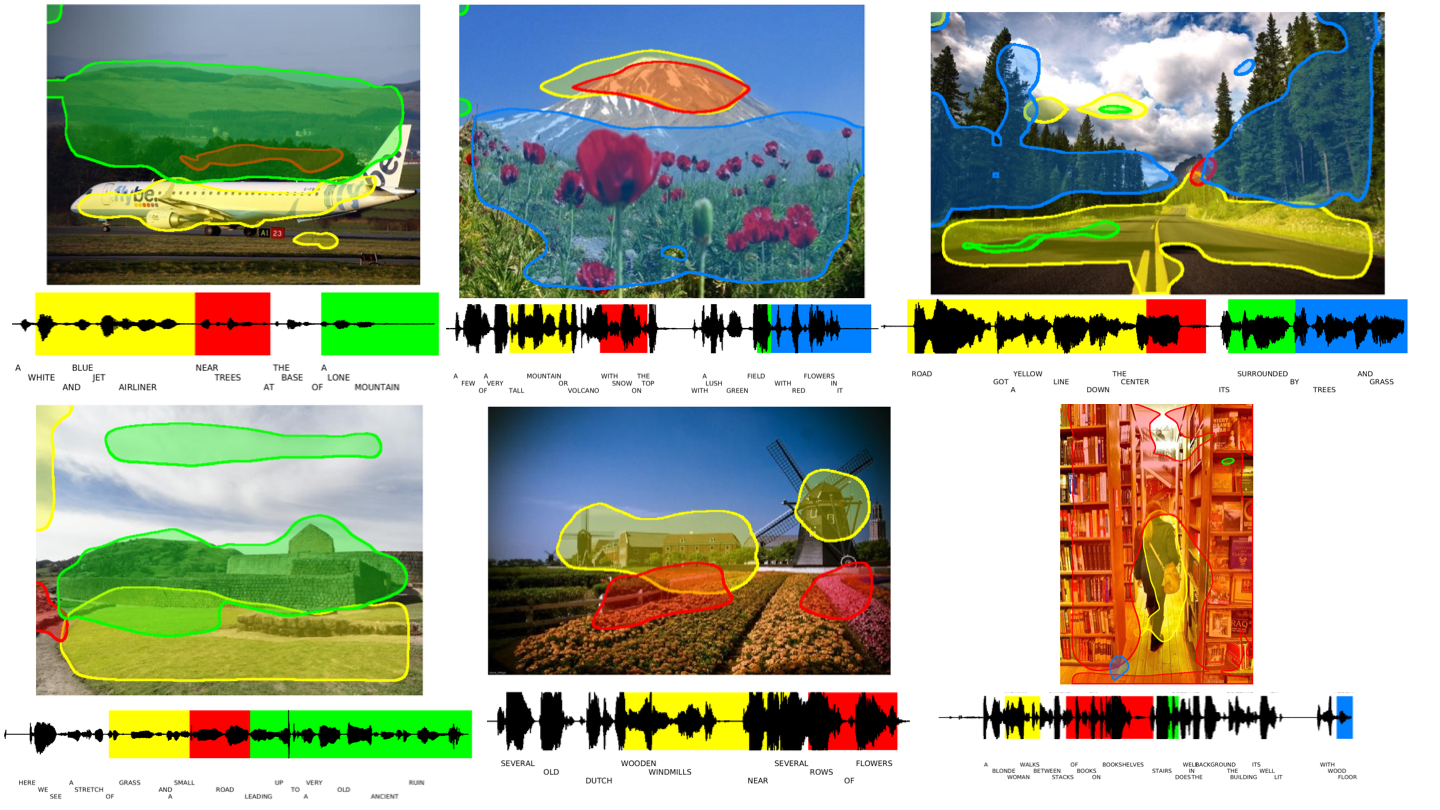}
\vspace{-.3in}
\caption{Additional examples of discovered image segments and speech fragments using the fully random MISA speech network. See supplementary materials for additional examples.}
    \label{fig:matchmapvis2}
\end{figure}

We can also extract volumetric connected components from the density and simultaneously project them down onto the image and spectrogram axes; visualizations of this are shown in Figures \ref{fig:matchmapvis} and \ref{fig:matchmapvis2}. For all visualizations, we found it necessary to apply a small amount of post-processing to the raw matchmaps in the form of thresholding and smoothing. The raw matchmaps can appear somewhat fragmented, so we first apply a sliding pooling window (max or average) with a size of 7 frames across the temporal dimension of the raw matchmap. Next, we normalize the matchmap scores to fall within the interval [0, 1] and sum to 1. Finally, we keep only the cells comprising the top $p$ percentage of the total mass within the matchmap, setting all others to zero. In practice, we found that $p$ values between 0.15 and 0.3 produced attractive results.

\section{Conclusions}

In this paper, we introduced audio-visual ``matchmap'' neural networks which are capable of directly learning the semantic correspondences between speech frames and image pixels without the need for annotated training data in either modality. We applied these networks for semantic image/spoken caption search, speech-prompted object localization, audio-visual clustering and concept discovery, and real-time, speech-driven, semantic highlighting. We also introduced an extended version of the Places audio caption dataset~\cite{harwath_nips}, doubling the total number of captions. Additionally, we introduced nearly 10,000 captions for the ADE20k dataset. There are numerous avenues for future work, including expansion of the models to handle videos, environmental sounds, additional languages, etc. It may possible to directly generate images given a spoken description, or generate artificial speech describing a visual scene. More focused datasets that go beyond simple spoken descriptions and explicitly address relations between objects within the scene could be leveraged to learn richer linguistic representations. Finally, a crucial element of human language learning is the dialog feedback loop, and future work should investigate the addition of that mechanism to the models.

\clearpage

{\small
\bibliographystyle{splncs}
\bibliography{egbib}
}

\end{document}